\documentclass{article}

\PassOptionsToPackage{numbers, compress}{natbib}

\usepackage[preprint]{neurips_2018}

\usepackage[utf8]{inputenc} 
\usepackage[T1]{fontenc}    
\usepackage{hyperref}       
\usepackage{url}            
\usepackage{booktabs}       
\usepackage{amsfonts}       
\usepackage{nicefrac}       
\usepackage{microtype}      
\usepackage{graphicx}
\usepackage{subfigure}

\title{PyramidBox++: High Performance Detector for Finding Tiny Face}

\author{%
  Zhihang~Li$^{1}$\thanks{Intern in Baidu},
  Xu~Tang$^{2}$,
  Junyu~Han$^{2}$,
  Jingtuo~Liu$^{2}$,
  and~Ran~He$^{1}$\\
  $^1$ CRIPAC \& NLPR, Institute of Automation, Chinese Academy of Sciences, Beijing, China. \\
  $^2$ Baidu, Inc. \\
  {\tt\small \{zhihang.li, rhe\}@nlpr.ia.ac.cn} \\
  {\tt \small \{tangxu02, hanjunyu, liujingtuo\}@baidu.com}
}

\begin{document}

\maketitle

\begin{abstract}
  With the rapid development of deep convolutional neural network, face detection has made great progress in recent years. WIDER FACE dataset, as a main benchmark, contributes greatly to this area. A large amount of methods have been put forward where PyramidBox designs an effective data augmentation strategy (Data-anchor-sampling) and context-based module for face detector. In this report, we improve each part to further boost the performance, including Balanced-data-anchor-sampling, Dual-PyramidAnchors and Dense Context Module. Specifically, Balanced-data-anchor-sampling obtains more uniform sampling of faces with different sizes. Dual-PyramidAnchors facilitate feature learning by introducing progressive anchor loss. Dense Context Module with dense connection not only enlarges receptive filed, but also passes information efficiently. Integrating these techniques, PyramidBox++ is constructed and achieves state-of-the-art performance in hard set. 
\end{abstract}

\section{Introduction}

Face detection, aiming at determining and locating the regions of faces in the natural images, is one of the fundamental steps in various face analysis, including face tracking \cite{kim2008face}, alignment \cite{xiong2013supervised,zhu2016face}, recognition \cite{parkhi2015deep,wu2018light}, synthesis \cite{huang2018introvae,wang2018face} etc. However, it is challenging to detect face accurately, because faces in the unconstrained environments have large intraclass variations, like large scale variance, occlusion, pose, illumination etc. Therefore, face detection has greatly raised and drawn much attention in computer vision.

So far great progress has been made in face detection. The seminal work Viola-Jones detector \cite{viola2004robust} is the first efficient detector, which adopts Haar-Like features to train a cascade of binary classifiers with AdaBoost algorithm. Since then, large numbers of subsequent works \cite{brubaker2008design,pham2007fast} have been proposed to boost the performance. In particular, the deformable part models (DPM) \cite{felzenszwalb2010object} are also introduced to model the different parts of face. This type of methods mainly depends on handcraft feature and carefully designing classifiers.

With the breakthrough of deep learning, image classification \cite{russakovsky2015imagenet} and object detection \cite{girshick2014rich,ren2015faster,liu2016ssd} have advanced considerably. As a special case of generic object detection, face detection also benefits from the CNN-based representation. Following the modern object detectors, state-of-the-art face detection algorithms can be roughly divided into two groups: two-stage face detectors (eg. R-CNN \cite{girshick2014rich}, Fast R-CNN \cite{girshick2015fast}, Faster R-CNN \cite{ren2015faster} etc.) and one-stage detectors (SSD \cite{liu2016ssd}, YOLO \cite{redmon2016you} etc). They have complementary merits and demerits: two-stage detector achieves the better performance but suffers from time-consuming inference, while one-stage detector has faster speed. Since face detection has the high demand of speed in real applications, the one-stage face detector attracts increasing attentions \cite{najibi2017ssh,zhang2017s,tang2018pyramidbox}.

Benefit from the development of image classification and object detection, we further improve Pyramidbox \cite{tang2018pyramidbox} including data augmentation, feature learning, context-aware prediction module and multi-task training. Specifically, a balanced-data-anchor-sampling is proposed to obtain a more uniform distribution among different scales of faces. Furthermore, we combine the PyramidAnchor \cite{tang2018pyramidbox} and dual shot structure \cite{li2018dsfd}, called Dual-PyramidAnchors, to make full use of context information. Inspired by DenseNet \cite{huang2017densely}, dense connection mechanism is utilized to pass information and gradient efficiently. In addition, multi-task training strategies, including segmentation and anchor free task, are employed to provide additional supervision. With integrating these tricks, we achieve state-of-the-art performance in hard set (small faces) of WIDER FACE.

\section{Related Work}

\textbf{Generic Object Detection.} Recently, object detection has been dominated by CNN-based detector. Following the milestone works of two-stage methods and one-stage methods, plenty of subsequent methods have been put forward to promote their developments. The cascade and refinement ideas are played vividly and incisively in both two-stage (cascade rcnn \cite{cai2018cascade}) and one-stage (RefineDet \cite{zhang2018single}) detectors. To strengthen the capacity of dealing with large scale variance in CNN-based detector, the series of SNIP \cite{singh2018analysis}, SNIPER \cite{singh2018sniper} and AutoFocus \cite{najibi2018autofocus} adopt a novel thought where they force detectors to focus on accurately detecting object in a certain range and expand the detection range by multi-scale testing. Thus they achieve the state-of-the-art performance on COCO dataset. Moreover, a quality assessment module (IoUNet \cite{jiang2018acquisition} and Scoring MaskRCNN \cite{Huang2019Mask}) is designed to rescore the predicted bboxes, whose goal is to solve the inconsistency of classification probability and bbox localization score. To purse high recall, it is necessary to tile massive dense anchors on high-resolution feature map. However, it results in an extreme imbalance of class that drastically impacts the classification task in detection \cite{lin2017focal}. An adaptive anchor tiling strategy, like MetaAnchor \cite{yang2018metaanchor} and Guided Anchor \cite{wang2019region}, is proposed to shrink search space efficiently.

\textbf{Face Detection.} Since the WIDER FACE dataset is built, large number of face detectors are proposed to locate faces under challenging environment, such as low- resolution imaging, tiny scale faces, large pose variations and occlusions in video surveillance. Wherein finding tiny faces is one of the research hotspots. S3FD \cite{zhang2017s} and \cite{zhu2018seeing} propose anchor matching strategy to improve the recall rate of tiny faces. Pyramidbox \cite{tang2018pyramidbox} fully exploits the context information to provide extra supervision for small faces. The super-resolution based on GAN \cite{bai2018finding} is introduced to face detection to make up the feature of low-resolution faces. Based on RefineDet \cite{zhang2018single}, SRN \cite{chi2018selective} investigates the effectiveness of cascade regression and classification on each level and find that two-step classification is used in shallow layers while two-step regression is used in deeper layers. DSFD \cite{li2018dsfd} improves several parts in Pyramidbox and achieves state-of-the-art performance.

\begin{figure}[t]
\centering
\includegraphics[width=0.9\linewidth]{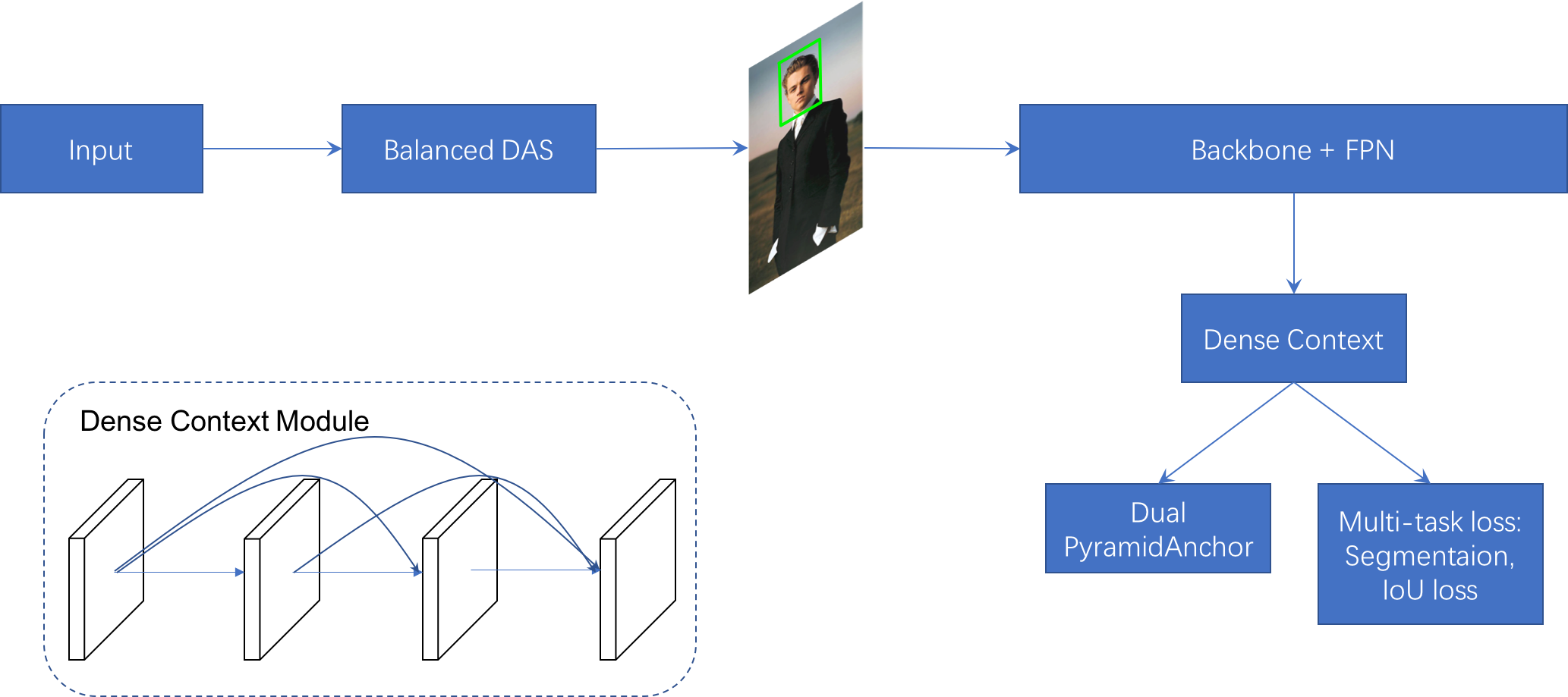}
  \caption{The brief overview of PyramidBox++. It consists Balanced-Data-anchor-sampling, Dense Context Module, Dual-PyramidAnchors and Multi-task loss. Particularly, the left-bottom part shows the detailed structure of Dense Context Module.}
  \label{fig:frame}
\end{figure}

\section{Method}
\subsection{Balanced-data-anchor-sampling}
We combine the original SSD-sampling \cite{liu2016ssd} and data-anchor-sampling (DAS) method \cite{tang2018pyramidbox} , where color distort, random crop and horizontal flip are done on the photo with a specified probability value. However, we find that DAS always introduce too many small faces, leading to the imbalance of face samples. Hence, we use a Balanced-data-anchor-sampling (BDAS) strategy. BDAS picks the anchor size with equal probability, and then the selected size will be acquired in the interval nearby the anchor size with equal probability too. Different from DAS, more face samples will be resized to bigger size with higher probability. Specifically, the face samples with the sizes ranging from 32 to 128 will count for a larger part for the whole face samples compared to DAS. In the implementation, we utilize BDAS with probability of 4/5 and SSD-sampling with probability of 1/5, respectively.

\subsection{Dual-PyramidAnchors}
For each target face, original PyramidAnchors \cite{tang2018pyramidbox} generate a series of anchors with larger regions centered in face to contain more contextual information, such as head, shoulder and body. PyramidAnchors choose the layers to set such anchors by matching the region size to the anchor size, which will supervise higher-level layers to learn more representable features. It is noteworthy that PyramidAnchors are implemented in a semi-supervised way under the assumption that regions with the same ratio and offset to different faces own similar contextual feature. In our Dual-PyramidAnchors, we introduce progressive anchor loss to PyramidAnchors referring to DSFD \cite{li2018dsfd} by setting some anchors to the features nearby the backbone, and it can help facilitate the features near the backbone. In prediction process, we only use output of the face branch in the second shot, so no additional computational cost is incurred at runtime.

\subsection{Dense Context Module}
Previous works, such as MSCNN \cite{cai2016unified}, SSH \cite{najibi2017ssh} and PyramidBox \cite{tang2018pyramidbox}, has demonstrated that delicately designing a predict module is effective for face detection. The underlying reason may be that receptive field is increased to cover more range of context information. However, too deep and complex predict module leads to difficulties in optimization and supervision. Inspired by dense connections in DenseNet \cite{huang2017densely}, we incorporate dense block into the predict module to pass information efficiently and preserve more multi-scale context feature. The detailed illustration shows in Figure \ref{fig:frame}. 

\subsection{Multi-task learning: segmentation and anchor free}
Multi-task learning has been proved effective in various computer vision tasks, which can help the network learn robust features. We make use of the task of segmentation \cite{wang2017face} and anchor free detection \cite{wang2018sface} to supervise the process of training. In the sub-task of segmentation, the segmentation branch is parallel to the classification branch and regression branch of detection in the head-architecture. Bounding box level segmentation ground truth is used for supervise our training process of segmentation, and different branch addressing different scales of the faces by anchor matching, the same as detection. The receptive field of the segmentation subnet is equivalent to the receptive field of detection subnet, which aims to ensure that both them concentrate on the same range of face scales. The segmentation branch introduced to our model will help to learn more discriminative features from face regions. Consequently, the classification subnet and the regression subnet in detection branch become easier, leading to better performance. Moreover, we introduce the anchor free detection branch. Just like yolo \cite{redmon2016you}, densebox\cite{huang2015densebox}, and unitbox\cite{yu2016unitbox}, this branch can directly acquire the bounding boxes without any anchors. 

\begin{figure*}[t]
\centering
\subfigure[Val: Easy]{
\label{fig:ve}
\includegraphics[width=0.49\linewidth]{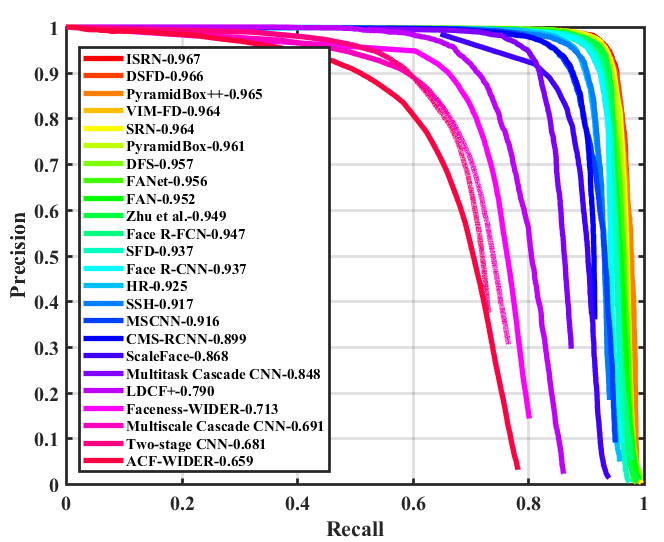}}
\subfigure[Test: Easy]{
\label{fig:te}
\includegraphics[width=0.49\linewidth]{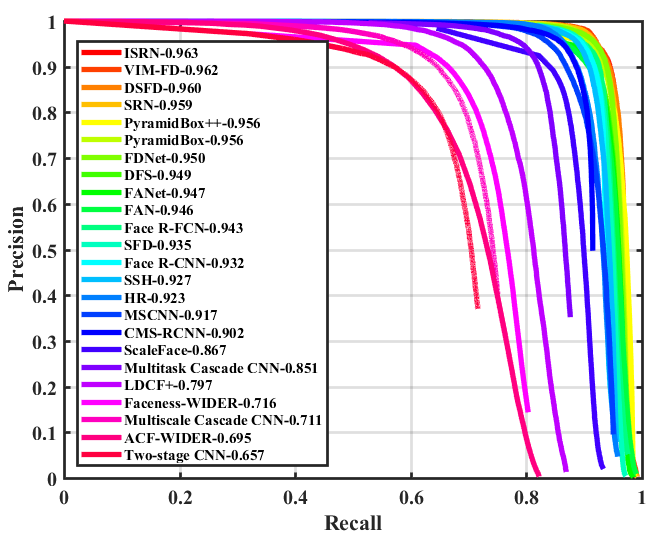}}
\subfigure[Val: Medium]{
\label{fig:vm}
\includegraphics[width=0.49\linewidth]{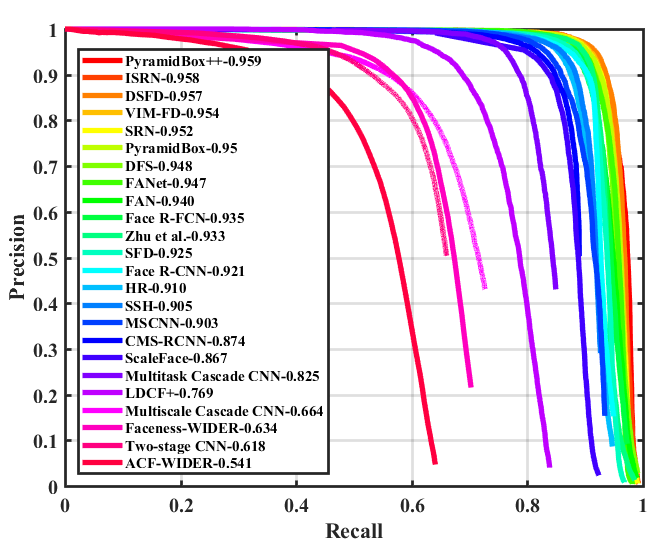}}
\subfigure[Test: Medium]{
\label{fig:tm}
\includegraphics[width=0.49\linewidth]{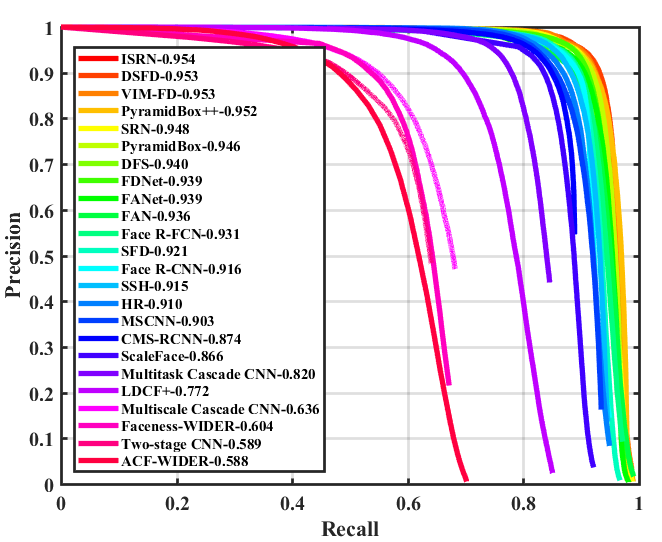}}
\subfigure[Val: Hard]{
\label{fig:vh}
\includegraphics[width=0.49\linewidth]{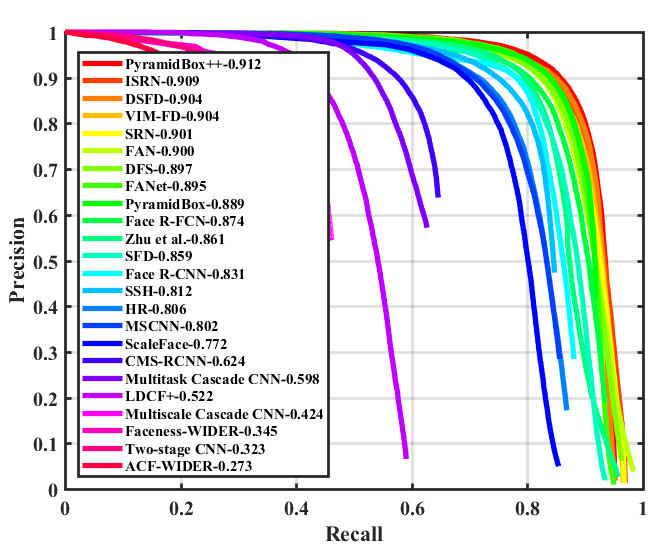}}
\subfigure[Test: Hard]{
\label{fig:th}
\includegraphics[width=0.49\linewidth]{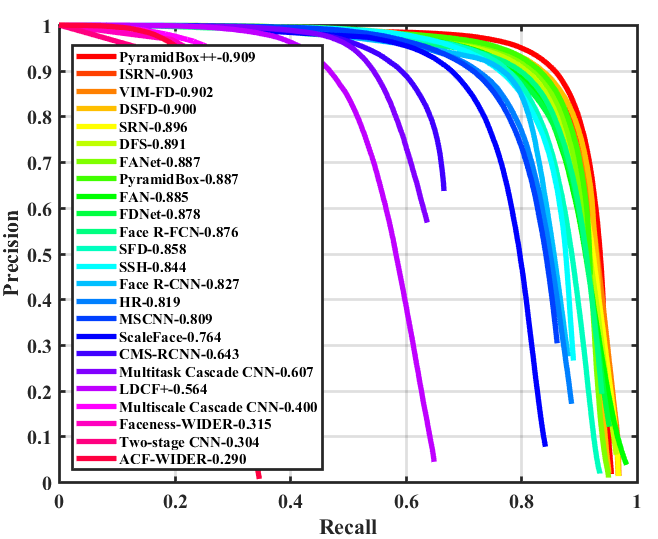}}
\vspace{-5mm}
\caption{Precision-recall curves on WIDER FACE validation and testing subsets.}
\vspace{-5mm}
\label{fig:wider-face-ap}
\end{figure*}

\section{Experiment}
\subsection{Implementation Detail}
We use resnet50 as a backbone network, which is initialized by pre-trained model in ImageNet \cite{he2016deep}. The newly additional layers are initialized with 'xavier'. We use mini-batch SGD with momentum of 0.9 and weight decay of 0.0005. The batch size is set to 28 on four GPUs. Warming up lr schedule is used in the first 3,000 iterations from 1e-6 to 4e-3, and decreases 10 times at iteration 80k and 100k, and the training ended at 120k iterations. It is noted that we use element-wise product in the low-level fpn \cite{lin2017feature} instead of element-wise summarization. Image warp is not used in data augmentation. Moreover, we filter out most of the simple negative samples using the negative threshold 0.99 to reduce the search space for the following operation\cite{chi2018selective}. Our method is based on Pytorch.

\subsection{Dataset}
WIDER FACE dataset \cite{yang2016wider}. It consists of 393,703 annotated face bounding boxes in 32, 203 images with variations in pose, scale, facial expression, occlusion, and lighting condition. The dataset is split into the training (40\%), validation (10\%) and testing (50\%) sets, and defines three levels of difficulty: Easy, Medium, Hard, based on the detection rate of EdgeBox \cite{zitnick2014edge}. Due to large scale variances and occlusion, WIDER FACE is the most challenging dataset in face detection. All the models are trained on the training set of the WIDER FACE dataset.

\subsection{Experimental Results}
\begin{figure}[h]
\centering
\includegraphics[width=1.0\textwidth]{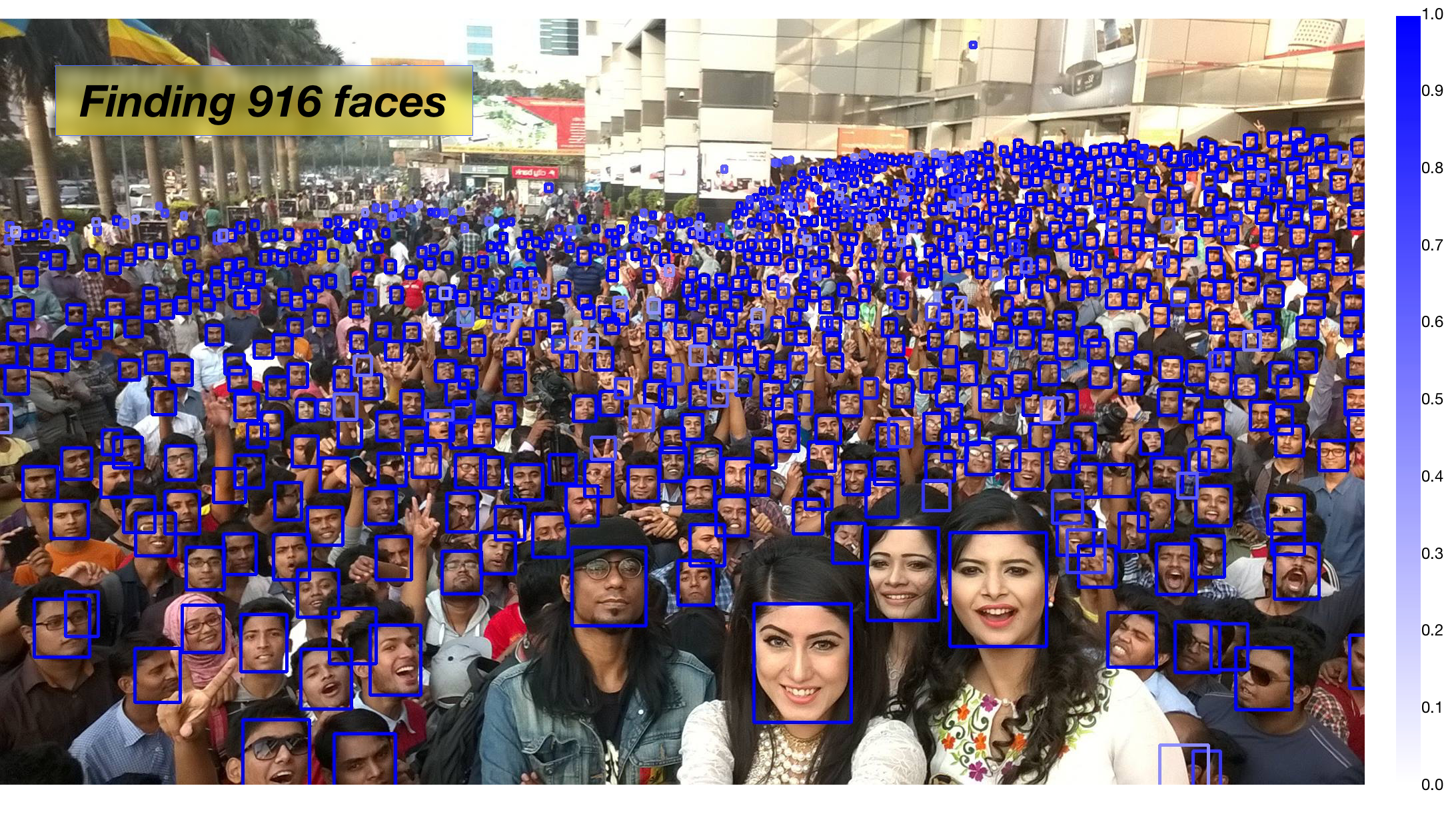}
\caption{Impressive qualitative result. VIM-FD finds $916$ faces out of the reported $1000$ faces. The confidences of the detections are presented in the color bar on the right hand. Best viewed in color.}
\label{fig:slumia}
\end{figure}

\begin{figure}[h]
\centering
\includegraphics[width=0.9\textwidth]{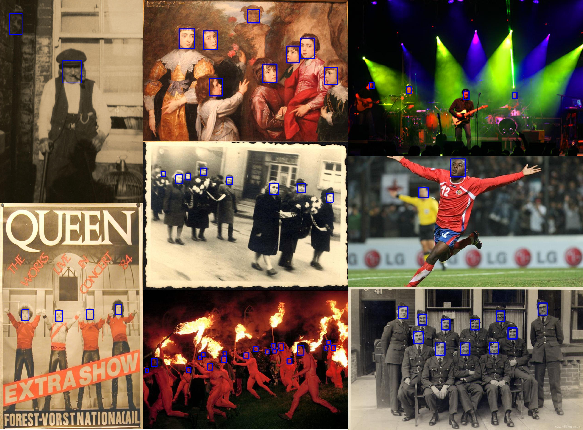}
\caption{The results of our PyramidBox++ across illumination and blur is shown in this figure, and blue represent the detector confidence above 0.8.}
\label{fig:illu_blur}
\end{figure}

\begin{figure}[!h]
\centering
\includegraphics[width=0.9\textwidth]{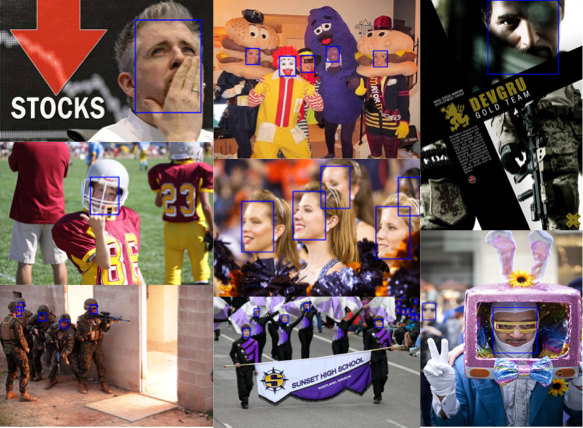}
\caption{The results of our PyramidBox++ across occlusion is shown in this figure, and blue represent the detector confidence above 0.8.}
\label{fig:occlusion}
\end{figure}

\begin{figure}[!h]
\centering
\includegraphics[width=0.9\textwidth]{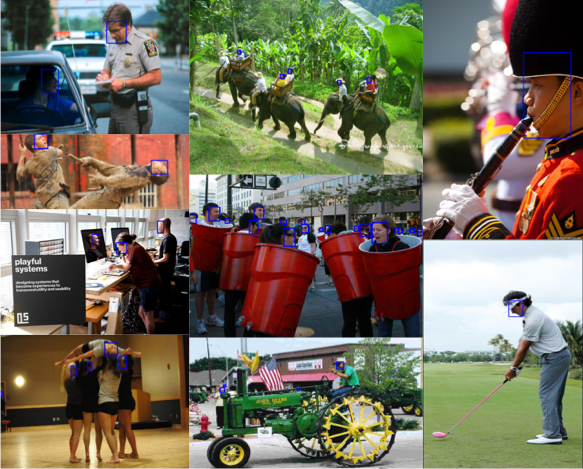}
\caption{The results of our PyramidBox++ across pose is shown in this figure, and blue represent the detector confidence above 0.8.}
\label{fig:pose}
\end{figure}

\begin{figure}[!h]
\centering
\includegraphics[width=0.9\textwidth]{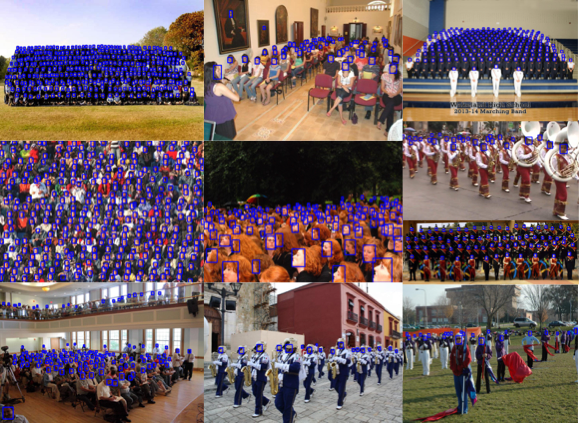}
\caption{The results of our PyramidBox++ across scale is shown in this figure, and blue represent the detector confidence above 0.8.}
\label{fig:scale}
\end{figure}

As shown in Figure \ref{fig:wider-face-ap}, we compare Pyramidbox++ with other state-of-the-art face detection methods on both validation and testing sets. The testing results are evaluated by the author. We find that Pyramidbox++ achieves comparable results against other state-of-the-art performance based on the average precision (AP) across the three subsets, i.e. 96.5\% (Easy), 95.9\% (Medium) and 91.2\% (Hard) for validation set, and 95.6\% (Easy), 95.2\% (Medium) and 90.9\% (Hard) for testing set. Especially on the hard subset which contains large amount of tiny faces, we outperform all approaches, which demonstrate the effectiveness to detect tiny faces. We also show a qualitative result of the World Largest Selfie in Figure \ref{fig:slumia}. Our detector can successfully detect 916 faces out of 1,000 faces. More experimental results, including scale, blur, expression, illumination, makeup, occlusion and pose, are shown in Figure \ref{fig:pose},\ref{fig:illu_blur},\ref{fig:occlusion},\ref{fig:scale}.

\section{Conclusion}
In this report, we exploit several tricks in recent works to further boost the performance of Pyramidbox, including Balanced-data-anchor-sampling, Dual-PyramidAnchors, Dense Context module, multitask training etc. Extensive experiments have been conducted on WIDER FACE dataset. Finally, the Pyramidbox++ achieves the state-of-the-art detection performance for tiny face on hard set.

\textbf{Acknowledgments.} We would like to thank Kang Du, Bin Dong and Shifeng Zhang for valuable discussions.

\bibliographystyle{unsrt}
\bibliography{egbib}

\end{document}